\documentclass{zapiski}
\usepackage{colortbl}
\usepackage[cp1251]{inputenc}

\usepackage[T2A]{fontenc}
\usepackage{mathptmx}
\usepackage{url}
\usepackage{latexsym}
\usepackage{multirow}
\usepackage{lipsum}
\usepackage{graphicx}
\usepackage{amssymb}
\usepackage{hyperref}
\usepackage{multirow}
\usepackage{booktabs}
\usepackage{tabularx}
\usepackage{caption}
\usepackage{cite}

\usepackage{paralist}
\usepackage{subcaption}
\usepackage{array}
\usepackage{setspace}

\usepackage{placeins}
\usepackage[table]{xcolor}
\usepackage{pgf}
\usepackage{pgfplots}
\usepackage{tikz}

\usepackage{amssymb}
\usepackage[table]{xcolor}
\usepackage{arydshln} 
\usepackage{graphicx} 
\usepackage{booktabs}  

\usepackage{array}
\newcolumntype{H}{>{\setbox0=\hbox\bgroup}c<{\egroup}@{}}
\newcolumntype{Z}{>{\setbox0=\hbox\bgroup}c<{\egroup}@{\hspace*{-\tabcolsep}}}

\usepackage{float}
\restylefloat{table}

\usepackage{todonotes}

\english

\begin{document}

\title[LLMs for SVG: A Review]{Leveraging Large Language Models For Scalable Vector Graphics Processing: A Review}

\author[B.~Malashenko]{Boris Malashenko}
\address[B.~Malashenko]{ITMO University, Kronverksky Pr. 49, St. Petersburg, Russia}
\email{btmalashenko@itmo.ru}

\author[I.~Jarsky]{Ivan Jarsky}
\address[I.~Jarsky]{ITMO University, Kronverksky Pr. 49, St. Petersburg, Russia}
\email{ivanjarsky@itmo.ru}

\author[V.~Efimova]{Valeria Efimova}
\address[V.~Efimova]{ITMO University, Kronverksky Pr. 49, St. Petersburg, Russia}
\email{vefimova@itmo.ru}

\begin{abstract}
In recent years, rapid advances in computer vision have significantly improved the processing and generation of raster images. 
However, vector graphics, which is essential in digital design, due to its scalability and ease of editing, have been relatively understudied.
Traditional vectorization techniques, which are often used in vector generation, suffer from long processing times and excessive output complexity, limiting their usability in practical applications. 
The advent of large language models (LLMs) has opened new possibilities for the generation, editing, and analysis of vector graphics, particularly in the SVG format, which is inherently text-based and well-suited for integration with LLMs.

This paper provides a systematic review of existing LLM-based approaches for SVG processing, categorizing them into three main tasks: generation, editing, and understanding. 
We observe notable models such as IconShop, StrokeNUWA, and StarVector, highlighting their strengths and limitations. 
Furthermore, we analyze benchmark datasets designed for assessing SVG-related tasks, including SVGEditBench, VGBench, and SGP-Bench, and conduct a series of experiments to evaluate various LLMs in these domains. 
Our results demonstrate that for vector graphics reasoning-enhanced models outperform standard LLMs, particularly in generation and understanding tasks. 
Furthermore, our findings underscore the need to develop more diverse and richly annotated datasets to further improve LLM capabilities in vector graphics tasks.
\end{abstract}

\keywords{Scalable Vector Graphics (SVGs) \and Large Language Models (LLMs) \and Image Generation.}

\maketitle

\section{Introduction}
\label{sec:intro}

In recent years, there has been a significant breakthrough in the field of computer vision, particularly in image generation and processing. 
Modern machine learning models~\cite{rombach2022high,saharia2022photorealistic,xu2022versatile,chang2024fluxfastsoftwarebasedcommunication} demonstrate impressive results when working with raster images, enabling advancements in graphic design, art, and data visualization.
At the same time, vector graphics are also widely used in digital design.
Vector graphics have a fundamentally different structure and offer several advantages, including scalability without quality loss, compact representation, and ease of editing.

Vector images are based on mathematical descriptions of shapes such as lines, curves, polygons, and text elements. 
Unlike raster images, which consist of pixels, vector images store shape information as a set of coordinates and instructions, making them ideal for use in design, web development, and printed media. 
The most popular vector graphics format is SVG (Scalable Vector Graphics), which is based on XML markup.

\begin{figure}
    \centering
    \includegraphics[width=0.9\linewidth]{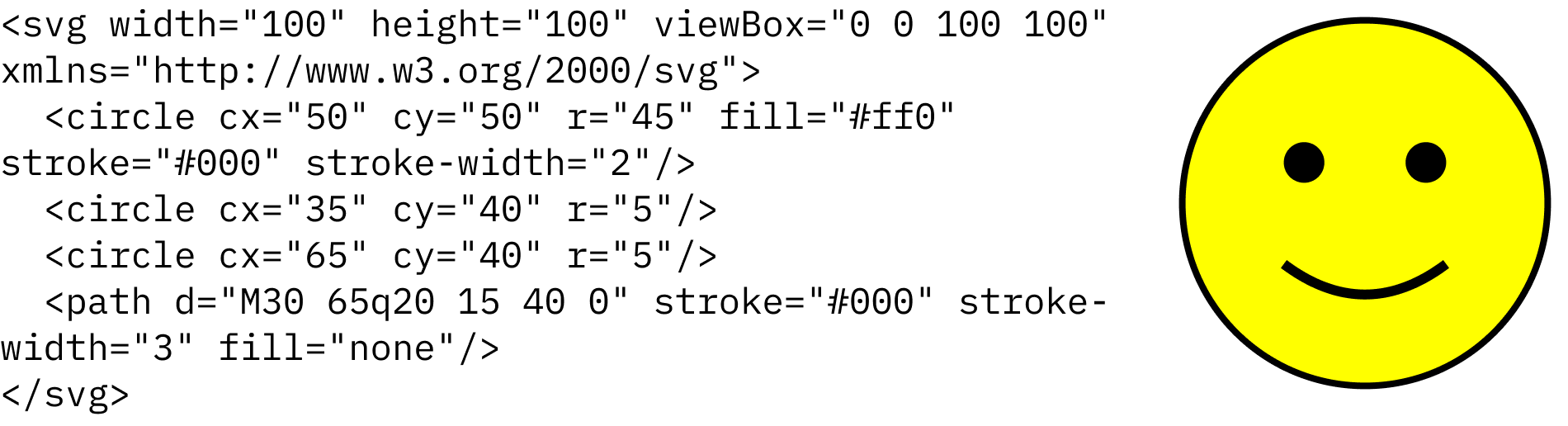}
    \caption{Example of smiley face in SVG format (left) and its rasterized analogue (right).}
    \label{fig:svg_example}
\end{figure}

Graphical elements in SVG files are text-based and are described using XML tags. For instance, a basic image in SVG format can be depicted as shown in Figure~\ref{fig:svg_example}. 
This approach allows easy editing using text editors and automated modifications using software tools. 
Moreover, SVG supports CSS styling and dynamic manipulation with JavaScript, making it particularly useful for web applications and interactive graphics.

Despite the clear value of vector graphics, the development of methods for their automatic generation and processing is still in its early stages.
The earliest pioneering models for vector graphics generation~\cite{svg-vae,deep-svg,covergan,reddy2021im2vec} produced relatively simple images that were often noisy and distorted.
However, their quality was insufficient for industrial use. 
A significant breakthrough in this area came with the introduction of DiffVG~\cite{li2020diffvg}, which enabled iterative vectorization of raster images with subsequent optimization of coordinates and other shape parameters. 
This led to the development of multiple optimization-based methods for image vectorization~\cite{ma2022towards,svgdreamer_xing_2023,livboc} and vector domain-specific style transfer methods~\cite{clipdraw,schaldenbrand2022styleclipdraw,vectornst}. 

Furthermore, the advent of diffusion models~\cite{rombach2022high,saharia2022photorealistic,xu2022versatile,mou2023t2iadapterlearningadaptersdig} has significantly improved the quality of generated raster images, facilitating substantial improvements in vector image generation through vectorization techniques~\cite{jain2022vectorfusion,svgdreamer_xing_2023}. 
The high fidelity of diffusion-generated images provides a robust foundation for vectorization processes, leading to more accurate and visually coherent vector outputs. 
Moreover, these advancements have enabled the optimization of vectorized representations through Score Distillation Sampling (SDS)~\cite{poole2022dreamfusiontextto3dusing2d} loss, which refines the generated vector images by leveraging the gradients of diffusion models, thereby improving their structural consistency and overall quality.

However, existing machine learning-based vectorization methods~\cite{li2020diffvg,ma2022towards,bazhenov2024evovec,bazhenov2024evolutionary,livboc} still face several serious challenges~\cite{dziuba2024image}.
At first, they often require more than $10$ minutes to process one image, especially when handling high-quality source images. 
Secondly, vectorization algorithms can generate an excessive number of primitives, making manual editing and post-processing utterly difficult. 
Additionally, traditional approaches do not allow designers to intuitively control the generation and editing process using text-based queries, limiting their practical applicability.

The study presented in~\cite{vectorweaver}, which focuses on generating SVGs, deliberately avoids utilizing DiffVG due to its inefficiency and frequent failures. Instead, the authors propose a novel generating approach based on a diffusion process. However, this process operates unconditionally and requires further investigation and refinement to enhance its effectiveness.

The emergence of large language models (LLMs)~\cite{achiam2023gpt,dubey2024llama,yang2024qwen2} has introduced new possibilities for vector graphics generation and processing, because vector files in SVG format are text-based. 
XML-based structure makes SVGs an ideal medium for interaction with language models, as LLMs can generate, analyze, and edit SVG code, enabling users to work with vector graphics using natural language commands. 
This approach simplifies the creation of complex graphic elements, reduces the effort of designers and developers, and fosters new techniques for automated image generation.

It is important to note that research on applying large language models to SVG has already started, with several studies demonstrating the potential of this approach. 
Specifically, some works~\cite{timofeenko2023vector,bubeck2023sparks,zou2024vgbench} showcase the feasibility of automatic SVG generation based on textual descriptions, as well as the application of LLMs for editing and optimizing existing vector images. 
These studies lay the groundwork for future technological advancements that combine the power of language models with the flexibility of vector formats.

We believe that, at present, language models are the only viable versatile approach for processing vector images. 
In this paper, we explore existing methods for generating and processing vector graphics using LLMs, analyze their advantages and limitations, and discuss the future prospects of this field.

Our contributions are as follows: 
\begin{itemize}
    \item We systematize approaches to vector graphics processing with LLMs.
    \item We verify and reassess prior research findings.
    \item We evaluate recently appeared LLMs, including novel reasoning models, for vector graphics tasks.
\end{itemize}


In this paper, we explore the role of large language models (LLMs) in vector graphics processing, highlighting their potential and drawbacks in SVG generation, editing, and understanding tasks. 
Section~\ref{sec:models} provides an overview of existing LLM-based models capable of SVG processing.
Section~\ref{sec:benchmarks} presents a detailed analysis of benchmark datasets and evaluation frameworks used to assess LLMs' capabilities in vector-related tasks, covering Image-text bridging, SVGEditBench~\cite{nishina2024svgeditbench}, SVG Taxonomy~\cite{xu2024exploring}, VGBench~\cite{zou2024vgbench}, and SGP-Bench~\cite{qiu2024can}. 
Section~\ref{sec:experiments} describes our experimental methodology, including evaluation metrics, implementation details, and the systematized results obtained in our study. 
Finally, Section~\ref{sec:conclusion} summarizes our findings, discusses limitations, and outlines future directions for research in this field.



\section{LLM-Based models capable of SVG processing}
\label{sec:models}
In this section, we present a review of existing approaches for vector graphic processing that leverage the capabilities of LLMs. 
Our focus is on (1) specialized models explicitly designed for vector graphics, (2) general-purpose LLMs. 
We exclude methods that rely on vectorization techniques, because while being effective in certain contexts, they are often computationally intensive and impractical for real-world applications. 
Additionally, this review does not cover models for text-to-image generation that do not utilize LLMs, since in this research we explore the current capabilities and potential improvements of LLMs in the domains of vector graphic generation, editing, and analysis -- areas where LLMs offer distinct advantages.


\subsection{IconShop}

One of the first models leveraging LLM for vector image generation is the IconShop model~\cite{wu2023iconshop}, which employs a Transformer Decoder-based autoregressive approach to generate vector icons by effectively capturing long-range dependencies within an SVG sequence. 
The model's architecture consists of three main modules: an SVG embedding module, which encodes vector icon sequences using a learnable embedding matrix; a text embedding module, leveraging pre-trained BERT~\cite{devlin2019bertpretrainingdeepbidirectional} tokenization to process input textual description; and a Transformer Decoder module, which finally generates text tokens of synthesized SVG image. 
The model operates within a constrained $100 \times 100$ bounding box, encoding SVG elements as tokenized sequences that include command types, coordinate arguments, and special markers. 
During training, text and SVG sequences are padded and concatenated together to a fixed length, allowing the autoregressive transformer to predict the next token based on previous inputs. 

The IconShop model was trained on the FIGR8 dataset~\cite{figr8}, which consists of black-and-white images paired with simple, often monosyllabic text labels. 
Although this dataset provides a suitable foundation for the initial exploration of vector image generation using LLMs, its simplicity also imposes certain limitations. 
The restricted complexity and diversity of the FIGR8 dataset hinder the model's ability to generate complex and varied images when faced with complex prompts. 
This highlights the need for more diverse and richly annotated datasets to fully utilize the potential of state-of-the-art models in the task of generating complex vector graphics.

\subsection{StrokeNUWA}

Another model connected to LLMs for vector image generation is the StrokeNUWA~\cite{tang2024strokenuwa} approach.
The authors adapted the successful autoregressive image generation approach based on VQ-VAE~\cite{oord2018neuraldiscreterepresentationlearning} to operating vector graphics. 
Similarly to raster images, StrokeNUWA utilizes VQ-VAE to convert vector images into a compressed representation as a sequence of tokens. 
Subsequently, an Encoder-Decoder Language Model generates vector images from textual descriptions. 
During training, the encoder remains frozen, meanwhile the decoder constructs its token vocabulary based on VQ-VAE and is responsible for the autoregressive generation of these tokens. 
The resulting token sequence is then decoded by the VQ-VAE decoder to reconstruct the final image.

Despite its novel approach, the results presented by the authors are limited to simple images composed of unfilled curved lines. 
This raises concerns about the model's applicability in practical settings, as its current capabilities appear significantly constrained.

\subsection{StarVector}


A recently introduced model, StarVector~\cite{rodriguez2023starvector}, enables the generation of vector images conditioned on textual prompts or input raster images. 
In this work, the authors present a Multi-Modal Large Language Model (MLLM) trained to process conditioning tokens derived from either an image encoder~\cite{dosovitskiy2021imageworth16x16words} or textual input. 
The model operates in an autoregressive manner, generating tokens that represent the output vector image based on the given conditioning input.

The study addresses three interrelated tasks: vector image generation from textual descriptions, raster-to-vector conversion, and diagram generation. 
To facilitate a comprehensive evaluation, the authors introduce multiple datasets of varying complexity. 
The assessment of model performance is conducted using their custom benchmarking framework, SVG-Bench~\cite{rodriguez2023starvector}.
For image vectorization, in addition to standard metrics such as MSE, SSIM, and LPIPS, the authors employ DinoScore~\cite{oquab2024dinov2learningrobustvisual} and further assess the number of generated tokens and conduct user studies. 
In the text-to-image generation task, the evaluation is based on FID, CLIP-FID, and CLIP Score metrics.

The reported results are promising, demonstrating the model's capability to produce high-quality vector images. 
However, despite its effectiveness, the generated images still exhibit noticeable artifacts, particularly when processing complex prompts. 
Moreover, reproducibility remains an issue, as the authors have not yet released the code, pre-trained models, benchmarks, or datasets.

\subsection{Empowering LLMs to Understand and Generate Complex Vector Graphics}


One of the most recent models for vector image generation and analysis using LLMs is SVG4LLM, introduced in~\cite{xing2024empowering}. 
In this work, the authors propose the new Vision-Language Large Model (VLLM) while introducing a key innovation: the integration of SVG-specific tokens into the shared token vocabulary. 
This enhancement enables the model to better capture the structural characteristics of the SVG code, leading to more accurate and coherent generation and analysis of vector graphics.

In addition, the study focuses on the challenge of annotating SVG images. 
To address this, the authors curated a specialized annotated dataset, which was utilized during training to improve the model's understanding of vector graphics. 
The reported results demonstrate strong performance, showcasing the potential of the model in both generation and analysis tasks. 
However, it is important to note that the dataset, source code, and trained model weights have not yet been released publicly. 
As a result, independent verification of the reported findings is currently impossible.

\subsection{General-Purpose LLMs}
Models, such as GPT-4~\cite{achiam2023gpt}, LLaMA~\cite{dubey2024llama}, and Qwen~\cite{yang2024qwen2}, have demonstrated impressive capabilities in a wide range of natural language processing (NLP) tasks, but their application to vector graphics remains underexplored. 
Unlike specialized models trained explicitly for SVG generation and manipulation, general-purpose LLMs possess broad adaptability and a vast pre-trained knowledge base, enabling them to reason about and process SVG code without domain-specific fine-tuning. 
Recent studies suggest that while these models can perform basic SVG-related tasks -- such as code generation, semantic understanding, and structural editing -- their effectiveness remains inconsistent, especially when dealing with intricate vector compositions. 
Moreover, the reliance on tokenization strategies optimized for textual data often leads to inefficiencies when encoding geometric primitives and SVG-specific commands. 
Despite these challenges, the flexibility of general-purpose LLMs could make them a valuable tool for exploratory research in SVG processing, particularly when combined with fine-tuning or retrieval-augmented generation techniques. 
As the capabilities of foundation models continue to expand, their role in vector graphics processing is likely to grow, driving the need for improved tokenization strategies and fine-grained task-specific adaptation.

\section{Benchmarks} \label{sec:benchmarks}

Scalable Vector Graphics (SVG), being a text-based format, aligns well with LLMs, allowing direct manipulation without rasterization. 
Although early studies~\cite{timofeenko2023vector, bubeck2023sparks} have shown that models like GPT can interpret and edit SVG, they mainly emphasize qualitative examples rather than systematic evaluation.
To truly assess LLMs' capabilities in processing vector graphics, standardized benchmarks and metrics are essential. 
This section examines recent efforts to establish such benchmarks.

\subsection{Image-text bridging}

\begin{figure}
    \centering
    \includegraphics[width=0.9\linewidth]{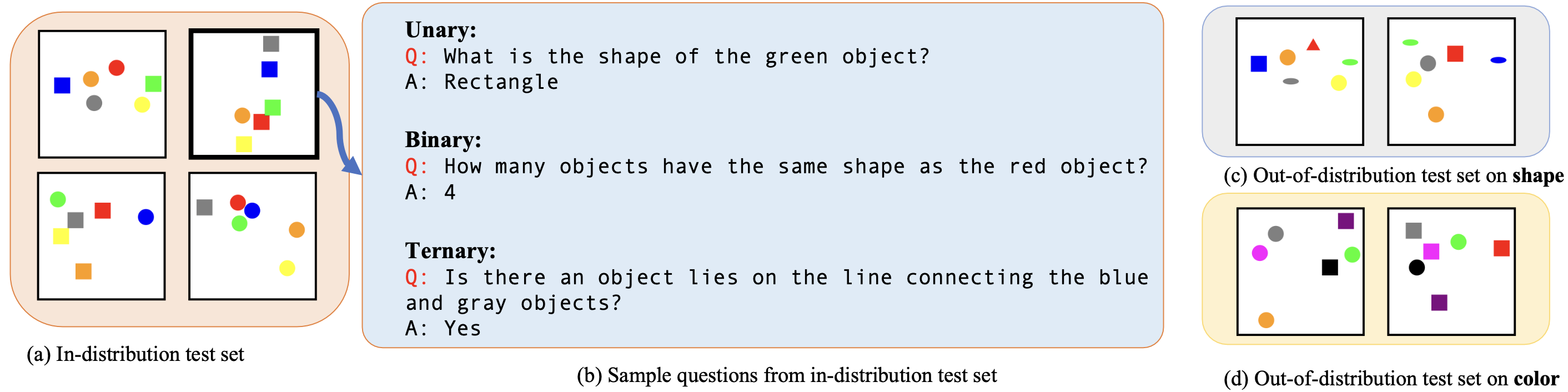}
    \caption{Image-text bridging~\cite{cai2024leveraginglargelanguagemodels} dataset curation. The figure is adapted from original paper.}
    \label{fig:image_text_bridge}
\end{figure}

The paper~\cite{cai2024leveraginglargelanguagemodels} presents the first systematic approach to evaluate LLMs on visual understanding tasks using SVG as a bridge between image and text data. 
To explore the visual reasoning capabilities of LLMs, the authors tested models on a range of tasks, including visual question answering, image classification under distribution shifts, few-shot learning, and image generation through visual prompting.
An example for one of the tasks is presented on Figure~\ref{fig:image_text_bridge}.

The results showed that GPT-4 performed moderate in understanding and working with SVG representations, especially in visual reasoning and answering questions. 
However, the models struggled with tasks requiring delicate visual details. 
While the study highlighted the potential of LLMs to handle a range of visual tasks without explicit visual encoders, it also exposed current limitations, suggesting that enhanced SVG representations may be needed for more complex visual reasoning.

\subsection{SVGEditBench}

\begin{figure}
    \centering
    \includegraphics[width=0.9\linewidth]{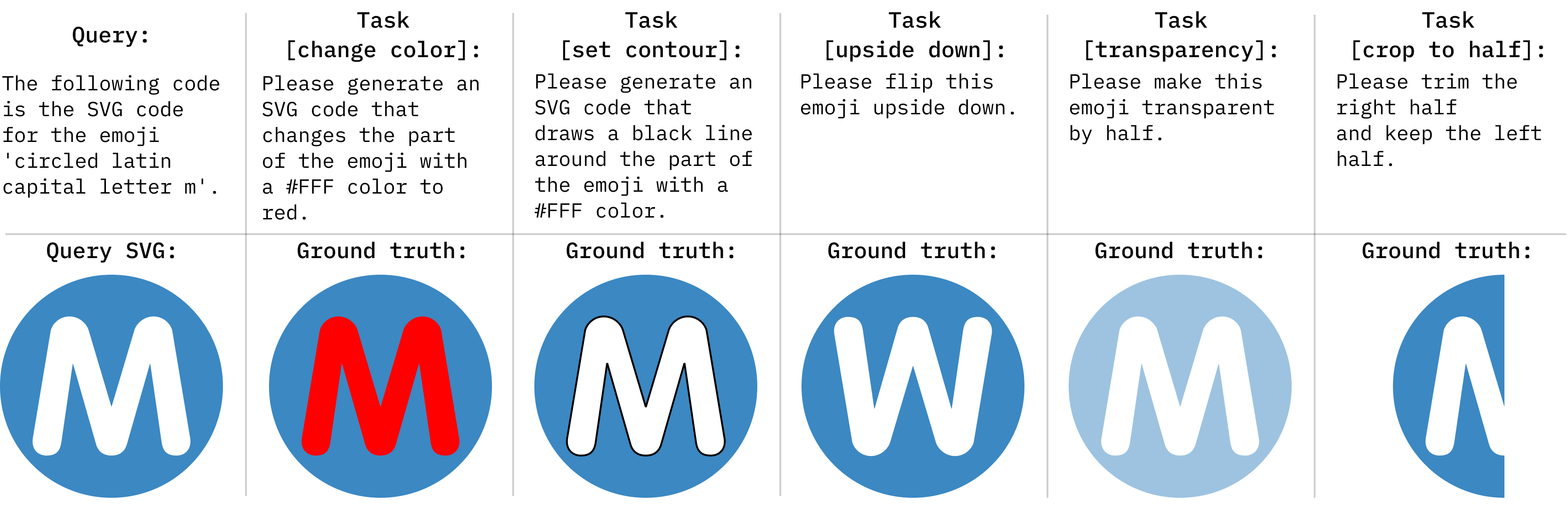}
    \caption{Example of all tasks except compression for a single sample from SVGEditBench~\cite{nishina2024svgeditbench}.}
    \label{fig:SVGEditBench_example}
\end{figure}

Another benchmark, called SVGEditBench~\cite{nishina2024svgeditbench}, is specifically designed to quantitatively evaluate LLMs abilities to edit SVG. 
The benchmark consists of six distinct editing tasks: changing colors, setting contours, cropping to half, rotating, applying transformations, and compressing SVG code, which allows a comprehensive evaluation of model performance (see examples in Figure~\ref{fig:SVGEditBench_example}).

By providing structured prompts and using clear quantitative metrics like Mean Squared Error (MSE) and compression ratios, the benchmark enables direct comparison in editing tasks.

The experimental results demonstrated that GPT-4 consistently outperformed GPT-3.5 across almost all tasks, showing superior accuracy in applying edits and understanding SVG structures. 
GPT-4 also displayed more sophisticated strategies, like optimizing code through intelligent smoothing in compression tasks. 
Despite these strengths, the study highlighted that even advanced models sometimes struggle with complex editing tasks, particularly when precision in SVG syntax is required.

\subsection{SVG Taxonomy}

Visual analytics is one of the SVG's possible applications, enabling the automation of tasks such as statistical analysis, data filtering, and other data workflow processes. 
SVG's code-based structure allows for complex visual manipulations, improving efficiency and accessibility in data exploration. 
The paper~\cite{xu2024exploring} examines how well LLMs can handle such tasks on SVG-based charts.

\begin{figure}
    \centering
    \includegraphics[width=0.97\linewidth]{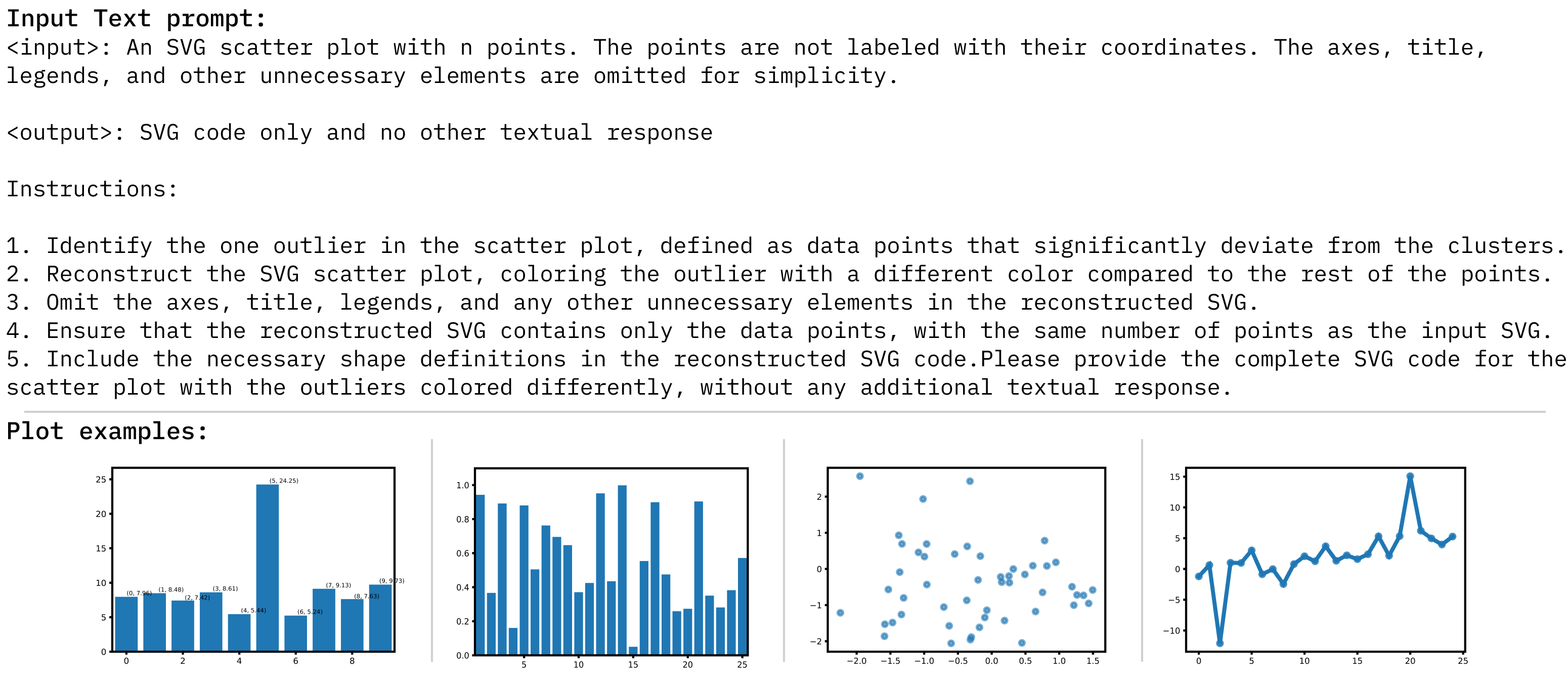}
    \caption{Example of a versatile prompt and various chart types from the SVG Taxonomy~\cite{xu2024exploring} benchmark.}
    \label{fig:taxomony_example}
\end{figure}

The study used GPT-4-turbo~\cite{achiam2023gpt} with zero-shot prompts to test $10$ low-level visual analytic tasks on SVG code. These tasks ranged from basic data extraction to more complex operations involving mathematical reasoning and pattern recognition. The model was also tested on sorting and identifying data distributions, as illustrated in Figure~\ref{fig:taxomony_example}, which presents examples of different chart types used in the evaluation.

The results showed that GPT-4-turbo excelled in pattern recognition tasks such as clustering and anomaly detection, but struggled with mathematical tasks, often generating inaccurate or hallucinated results, especially when the charts lacked value labels. 
Increasing data points amount did not always enhance performance, in certain cases improving numerical accuracy but also increasing the errors number. 
The study highlights LLMs' potential and limits in visual analytics, suggesting hybrid methods could improve the results, while better prompts and fine-tuning might overcome current weaknesses.

\subsection{VGBench}

The authors of the work~\cite{zou2024vgbench} took a step further by not only evaluating LLMs' ability to understand or edit vector graphics but generate. 
In the VGBench, they introduced two key components for comprehensive model assessment: Vector Graphics Question Answering (VGQA) for evaluating vector graphics understanding and Vector Generation (VGen) for assessing generation capabilities.

\begin{figure}
    \centering
    \includegraphics[width=0.9\linewidth]{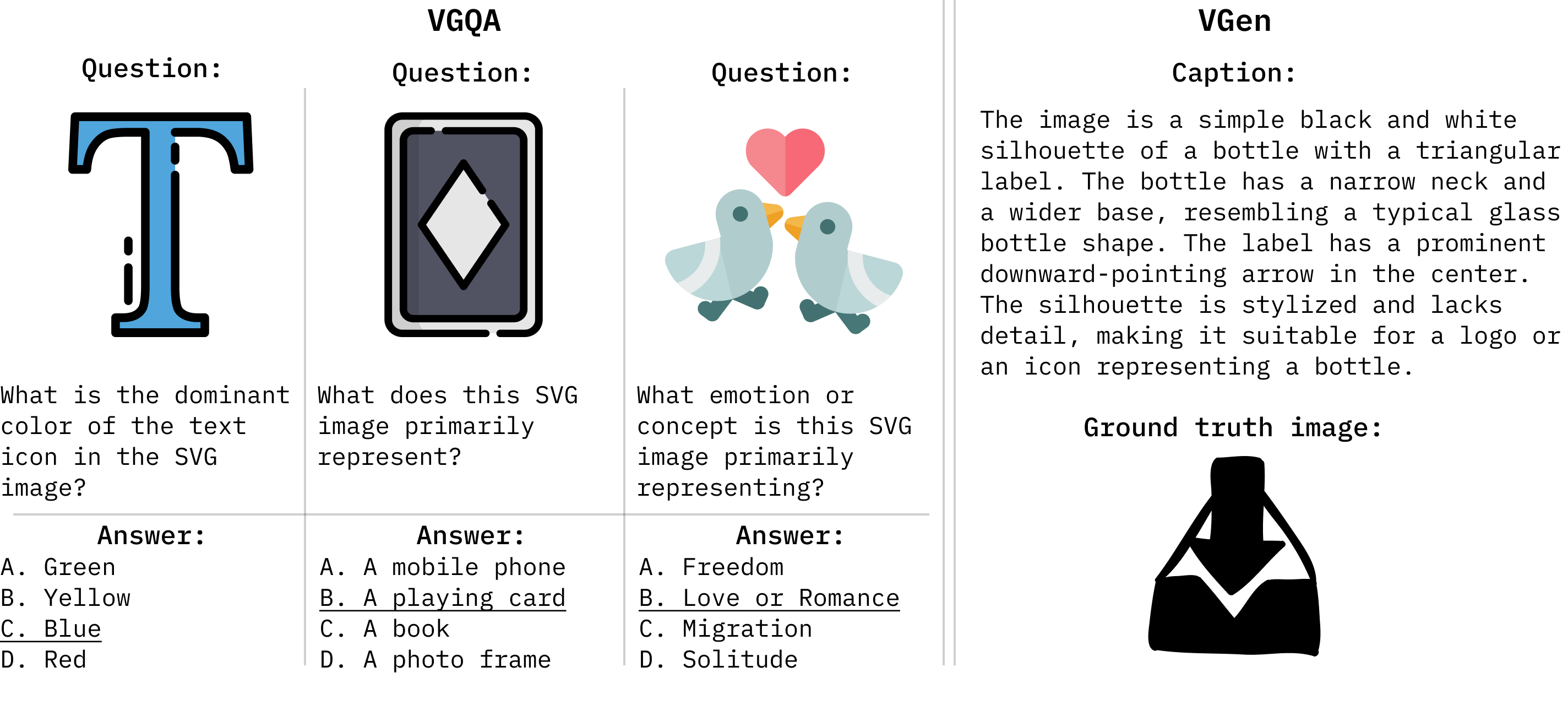}
    \caption{Examples of VGQA tasks for color, category, and usage, along with a sample from VGen -- both subtasks of VGBench. Ground truth answers are highlighted.}
    \label{fig:vgbench_example}
\end{figure}

VGQA focuses on tasks like identifying colors, categories, and usage of depicted subjects across different vector formats such as SVG, TikZ~\cite{tantau:2013a}, and Graphviz~\cite{ellson2002graphviz}. 
A semi-automated pipeline was used to create question-answer pairs. Based on rasterized SVG images, GPT-4V~\cite{zhang2023gpt} generated the initial dataset, which was later filtered out by human annotators to ensure quality. 
This setup allowed the authors to thoroughly evaluate how well LLMs can interpret various aspects of vector graphics, from basic features to complex semantic relationships, as illustrated in Figure~\ref{fig:vgbench_example}.

The VGen assessed the models' ability to generate vector graphics from textual descriptions. 
GPT-4V first produced captions for existing images, and LLMs were then asked for generating corresponding vector graphics code based on these descriptions. 
The quality of the generated graphics was evaluated using CLIP Score and Fr\'echet Inception Distance (FID), measuring both semantic alignment with the text and visual similarity to the originals. 
Results showed that LLMs, particularly GPT-4, demonstrated capabilities in both understanding and generating vector graphics, especially for high-level semantic formats such as TikZ and Graphviz. 
In contrast, a low-level format like SVG, based on geometric primitives, is a greater challenge for the models.

\subsection{SGP-Bench}

\begin{figure}
    \centering
    \includegraphics[width=0.9\linewidth]{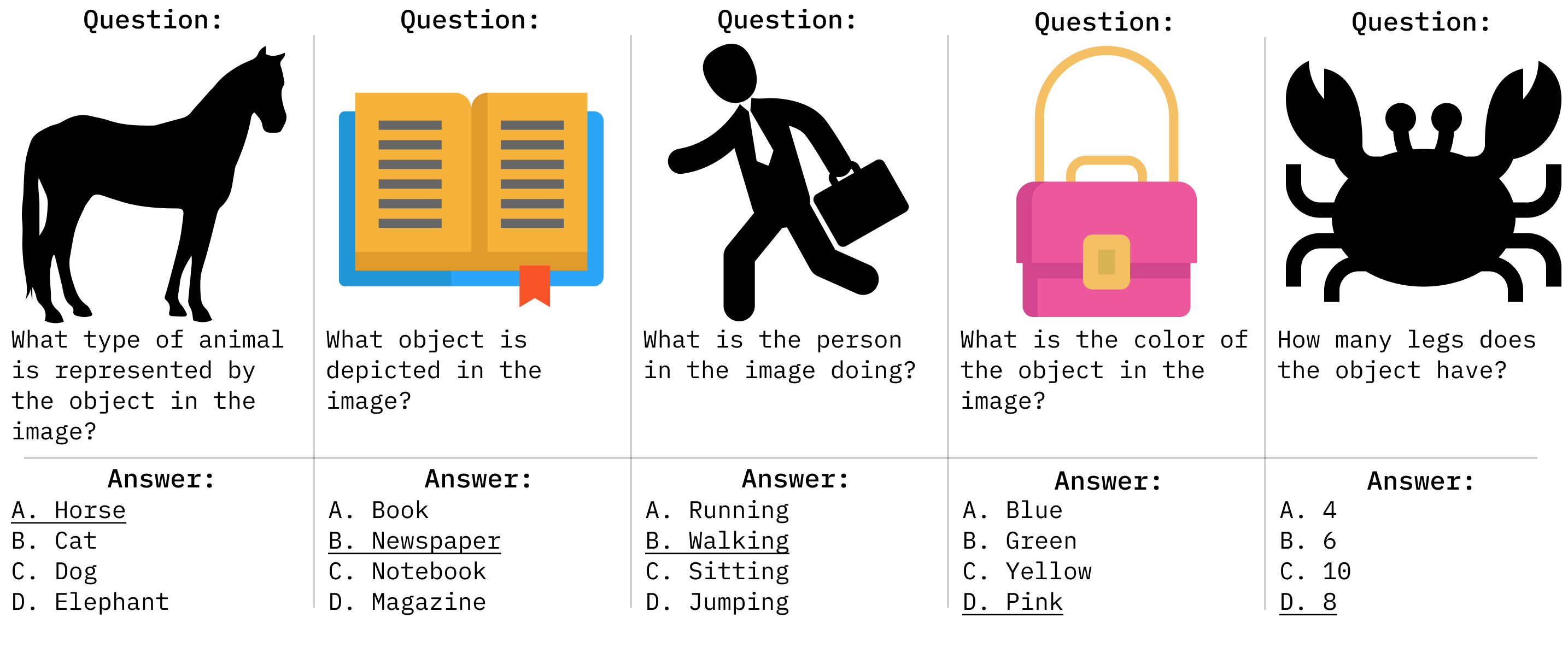}
    \caption{Random samples from SGP-Bench~\cite{qiu2024can} with highlighted ground truth answers. Notably, the second, third, and last examples are ambiguous.}
    \label{fig:sgp_bench_example}
\end{figure}

In the paper~\cite{qiu2024can}, the authors follow a novel approach by not only evaluating LLMs' abilities to understand vector graphics code (symbolic graphics programs, SGP), but also exploring how to enhance these
capabilities.
They introduced SGP-Bench, which evaluates two main aspects: (1) semantic understanding, where models answer questions about images without seeing them (illustrated in Figure~\ref{fig:sgp_bench_example}), and (2) semantic consistency, which tests how robustly LLMs interpret programs when compiled objects undergo transformations like rotations or translations.
Moreover, the authors proposed Symbolic Instruction Tuning (SIT), a fine-tuning method using instruction-based data derived from symbolic graphics programs. The SIT technique is aimed to improve LLMs' understanding of structured data.

The results showed that LLM demonstrates abilities to interpret symbolic graphics. However, significant limitations remain, particularly in tasks requiring complex spatial reasoning or understanding fine-grained details. 
Large proprietary models consistently outperform smaller open-source models, reflecting a clear scaling effect. 
SIT notably enhanced model performance, not only improving understanding of symbolic programs, but also boosting general reasoning abilities across other benchmarks.

\subsection{Summary}

To fully utilize the potential of vector graphic processing benchmarks, we can aggregate all tasks from these benchmarks and categorize them into three main groups: generation, editing, and recognition. 
The Table~\ref{table:bench_comp} summarizes the number of tasks related to SVG in each category across different benchmarks, providing an overview of their distribution and focus areas.
As can be seen from the table, no single benchmark comprehensively covers all three task categories, we recommend using two or more benchmarks to fully evaluate the capability of LLMs to work with SVG data.

\begin{table}
\centering
\footnotesize
\begin{tabular}{lccccc}
\hline
 Benchmark                   & Generation & Editing & Understanding \\ \hline
Image-Text bridging\cite{cai2024leveraginglargelanguagemodels} & -          & -       & 3             \\
SVGEditBench\cite{nishina2024svgeditbench}       & -          & 6       & -           \\
SVG Taxonomy~\cite{xu2024exploring}        & -          & 5       & 5          & &   \\
VGBench~\cite{zou2024vgbench}           & 2          & -       & 3           \\
SGP-Bench~\cite{qiu2024can}          & -          & -       & 5          \\ \hline
\end{tabular}
\caption{SVG-related task aggregation by type across benchmarks.}
\label{table:bench_comp}
\end{table}


Our analysis reveals that a significant portion of samples in both VGBench and SGP-Bench are inaccurate due to synthetic data generation without thorough validation. 
This highlights the need for human-based filtering to ensure data quality, particularly for such a complex task.
Another common limitation across all benchmarks is the relatively low complexity of the SVG files in the datasets. 
In real-world scenarios, SVG images often contain $500$ or even $1000$ paths, whereas the benchmark datasets are significantly simpler. 
To illustrate this, we present the Table~\ref{table:statistics}, which contains additional SVG-related statistics for benchmarks where initial datasets are available. 
Specifically, we compute the average number of nodes (SVG tags), figures, paths, and points. 
Here, "points" refer to individual numerical values rather than coordinate pairs, as some points are represented by a shift along a single axis and encoded as a single number. 
Points are counted for paths, lines, polygons, and polylines.


\begin{table}
\centering
\begin{tabular}{lcccc}
\hline
\multirow{2}{*}{Benchmark} & \multicolumn{4}{c}{Average statistics} \\ \cline{2-5} 
                           & Nodes   & Figures   & Paths  & Points  \\ \hline
SVGEditBench               & 8.89    & 7.82      & 6.12   & 337.22  \\
SVG Taxonomy               & 26.85   & 7.96      & 3.98   & 55.58   \\
VGBench                    & 25.45   & 7.76      & 5.64   & 414.76  \\
SGP-Bench                  & 11.83   & 7.81      & 5.50   & 334.38  \\ \hline
\end{tabular}
\caption{Average number of nodes (SVG tags), figures, paths, and points for each available benchmark.}
\label{table:statistics}
\end{table}

\section{Experiments}\label{sec:experiments}

In this section, we apply the methods using the methods with code available on benchmarks, which include the original datasets and evaluation scripts. 
Additionally, we assess a diverse set of LLMs to evaluate their performance across these benchmarks.

\subsection{Metrics}

Each benchmark includes its own specific metrics. 
However, to ensure consistency and enable the aggregation of scores across different benchmarks, we normalize these metrics. 
Additionally, since we evaluate multiple LLMs on a large set of tasks, we subsample the original datasets to reduce their size and computational budget.

\subsubsection{SVGEditBench} There are $6$ tasks in this benchmark, each evaluated using the MSE as the base metric. 
Calculated MSE values can be small and vary across tasks, we normalize all metrics based on the ground truth values presented in the original paper.

For each task $t$ and model $m$, let $s^m_t$ denote the score achieved, and $s^{0}_t$ represent the baseline MSE when no edits are applied. 
We rescale the MSE for each task to ensure values range from $0$ to $1$. 
We calculate the normalized score $\hat{s}_t$ as in Equation~\ref{eq:svgeditbench_norm}. The resulting benchmark score is the average across the six tasks.

\begin{equation}
\hat{s}_t = \begin{cases}
    \sqrt{\frac{s_t^0 - s_t}{s_t^0}},& s_t < s_t^0; \\
    0,& \text{otherwise}.
\end{cases}
\label{eq:svgeditbench_norm}
\end{equation}

For each task, we select the first $20$ images from the original dataset and evaluate all models on this identical set, resulting in a total of $120$ runs per model. 
If a model fails to produce a valid SVG output, the corresponding sample is excluded from the final calculations.

\subsubsection{VGBench} For VGen we randomly select $100$ samples from the initial dataset. 
Originally, two metrics are computed: CLIP Score and FID. 
However, we omit the FID results, as our observations, along with findings from the original paper, show that the metric is highly noisy and unreliable in this context. 
To ensure consistency, we normalize the CLIP Score values using the ground truth provided in the original paper -- $0.256$. 
To further reduce the impact of noise from the CLIP evaluator, we bound the CLIP Score at $1.0$ if the evaluated value surpasses the ground truth. 
The normalized CLIP Score for model $m$ is defined in Equation~\ref{eq:clip}, where $CLIP_0$ is ground truth and $CLIP_m$ is the Long-CLIP~\cite{zhang2024long} output for the model $m$.

\begin{equation}
\hat{CLIP_m} = 
\begin{cases}
    1 - \frac{CLIP_0 - CLIP_m}{CLIP_0}, & CLIP_m \leqslant CLIP_0;\\
    1, & \text{otherwise}.
\end{cases}
\label{eq:clip}
\end{equation}

VGQA comprises three subtasks: color, category, and object functionality recognition. 
For each subtask, we use the first $50$ images from the original dataset. 
We retain the original evaluation metric, which is calculated as an accuracy score. 
The final score is computed as the average accuracy across all three subtasks.

\subsubsection{SGP} The benchmark features five tasks centered on SVG comprehension: semantics, color, count, shape, and reasoning. 
We selected a $10\%$ subset of the original dataset ($434$ images) and evaluated performance across these tasks. 
The average scores were computed following the approach outlined in the original paper, using the same accuracy-based metric as in VGQA.

\subsection{Experimental Setup}

For all our experiments, we utilize original author code from repositories. 
We leverage openrouter.ai~\footnote{openrouter.ai} service
as a simple way to access multiple models. 
To ensure the results are reproducible, the generation temperature is fixed at $0$, other parameters left unchanged; system prompt is not specified.

\subsection{Results} For each of the metrics discussed above, we evaluated a range of models. 
Unfortunately, among the LLM-based methods, only IconShop provides both code and weights for evaluation. 
The results are presented in Table~\ref{table:main_comparision}.

\begin{table}
\centering
\renewcommand{\arraystretch}{1.1}
\setlength{\tabcolsep}{2pt}
\footnotesize
\resizebox{\textwidth}{!}{%
\begin{tabular}{|l|c c c c c|c:c:c:c|}
\hline
\multirow{2}{*}{Model} & \cellcolor[HTML]{FFCCCC}Generation & \cellcolor[HTML]{CCFFCC}Editing & \multicolumn{2}{c}{\cellcolor[HTML]{CCCCFF}Understanding} & \multirow{2}{*}{Avg} & \multirow{2}{*}{O} & \multirow{2}{*}{R} & \multirow{2}{*}{C} & \multirow{2}{*}{MoE} \\ \cline{2-5}
 & \cellcolor[HTML]{FFCCCC}VGen & \cellcolor[HTML]{CCFFCC}SVGEdit & \cellcolor[HTML]{CCCCFF}VGQA & \cellcolor[HTML]{CCCCFF}SGP & & & & & \\
\hline
IconShop & 0.816 & - & - & - & - & \checkmark & & & \\
Codestral-2501 & 0.889 & 0.901 & 0.273 & 0.484 & 0.723 & & & \checkmark & \\
Deepseek-R1 & 0.957 & \underline{0.926} & 0.587 & 0.710 & 0.844 & \checkmark & \checkmark & & \checkmark \\
Deepseek-R1-LLama-70B & 0.861 & 0.751 & 0.387 & 0.521 & 0.689 & \checkmark & \checkmark & & \\
Deepseek-R1-Qwen-1.5B & $\varnothing$ & 0.109 & 0.067 & 0.134 & - & \checkmark & \checkmark & & \\
Deepseek-R1-Qwen-32B & 0.858 & 0.721 & 0.407 & 0.401 & 0.661 & \checkmark & \checkmark & & \\
Deepseek-v3 & 0.923 & 0.642 & 0.460 & 0.606 & 0.699 & \checkmark & & & \checkmark \\
Gemma-2-9B & 0.857 & 0.084 & 0.060 & 0.429 & 0.395 & \checkmark & & & \\
Gemma-3-27B-it & 0.898 & 0.605 & 0.390 & 0.565 & 0.660 & \checkmark & & & \\
gemini-2.5-flash-preview & 0.940 & 0.899 & 0.680 & 0.459 & 0.803 & & & & \\
gemini-2.5-pro-preview & \underline{0.961} & 0.882 & \underline{0.727} & \underline{0.765} & \underline{0.863} & & & & \\
GPT-4.1 & 0.960 & 0.869 & 0.593 & 0.721 & 0.829 & & \checkmark & \checkmark & \\
GPT-4.1-mini & 0.944 & 0.849 & 0.560 & 0.671 & 0.803 & & \checkmark & \checkmark & \\
GPT-4.1-nano & 0.925 & 0.591 & 0.293 & 0.516 & 0.640 & & \checkmark & \checkmark & \\
GPT-4o & 0.947 & 0.827 & 0.653 & 0.670 & 0.812 & & & & \\
GPT-4o-mini & 0.910 & 0.727 & 0.380 & 0.574 & 0.707 & & & & \\
GPT-o4-mini & 0.949 & 0.914 & 0.700 & 0.696 & 0.857 & & \checkmark & \checkmark & \\
LLama3.2-1B & 0.805 & 0.073 & 0.220 & 0.168 & 0.357 & \checkmark & & & \\
LLama3.2-3B & 0.834 & 0.063 & 0.193 & 0.406 & 0.399 & \checkmark & & & \\
LLama3.3-70B & 0.863 & 0.460 & 0.347 & 0.558 & 0.592 & \checkmark & & & \\
LLama4-Maverick & 0.911 & 0.805 & 0.450 & 0.627 & 0.752 & \checkmark & & & \checkmark \\
Mistral-NeMo-12B & 0.848 & 0.267 & 0.253 & 0.419 & 0.484 & \checkmark & & & \\
Mistral-Small-24B & 0.866 & 0.722 & 0.187 & 0.491 & 0.642 & \checkmark & & & \\
Qwen-2.5-7B & 0.865 & 0.081 & 0.367 & 0.475 & 0.456 & \checkmark & & & \\
Qwen-2.5-72B & 0.872 & 0.533 & 0.372 & 0.567 & 0.625 & \checkmark & & & \\
Qwen-2.5-Coder-32B & 0.890 & 0.545 & 0.333 & 0.532 & 0.623 & \checkmark & & \checkmark & \\
Qwen-3-235B-A22B & 0.927 & 0.818 & 0.500 & 0.567 & 0.760 & \checkmark & & & \checkmark \\
Qwen-3-30B-A3B & 0.897 & 0.833 & 0.380 & 0.500 & 0.723 & \checkmark & & & \checkmark \\
Qwen-3-32B & 0.901 & 0.792 & 0.460 & 0.530 & 0.729 & \checkmark & & & \\
Qwen-3-8B & 0.889 & 0.419 & 0.450 & 0.435 & 0.584 & \checkmark & & & \\
Qwen-Max & 0.906 & 0.599 & 0.440 & 0.528 & 0.663 & & & & \\
Qwen-QwQ-32B & 0.905 & 0.651 & 0.400 & 0.535 & 0.675 & \checkmark & \checkmark & &  \\
Qwen-Turbo & 0.858 & 0.574 & 0.296 & 0.477 & 0.606 & &  & & \\
\hline
\end{tabular}%
}
\caption{Performance comparison of various models across benchmarks. Last columns show model attributes: O (open-source), R (reasoning), C (code-focused), MoE (Mixture-of-Experts).}
\label{table:main_comparision}
\end{table}

\begin{figure}
    \centering 
    \includegraphics[width=1\linewidth]{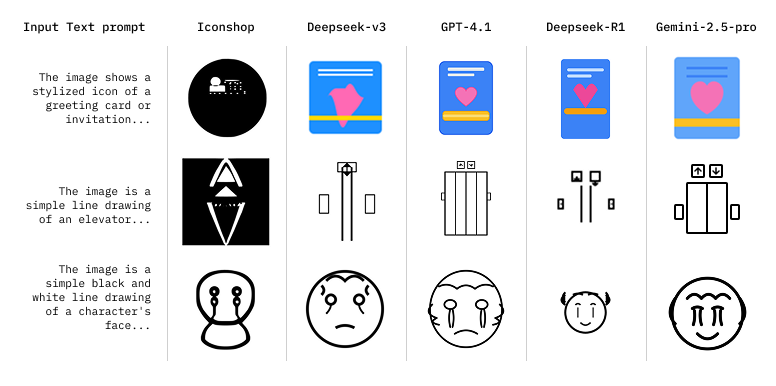} 
    \caption{Randomly selected images generated by various models.} 
    \label{fig:models_generation} 
\end{figure}

\begin{figure}
    \centering
    \includegraphics[width=0.9\linewidth]{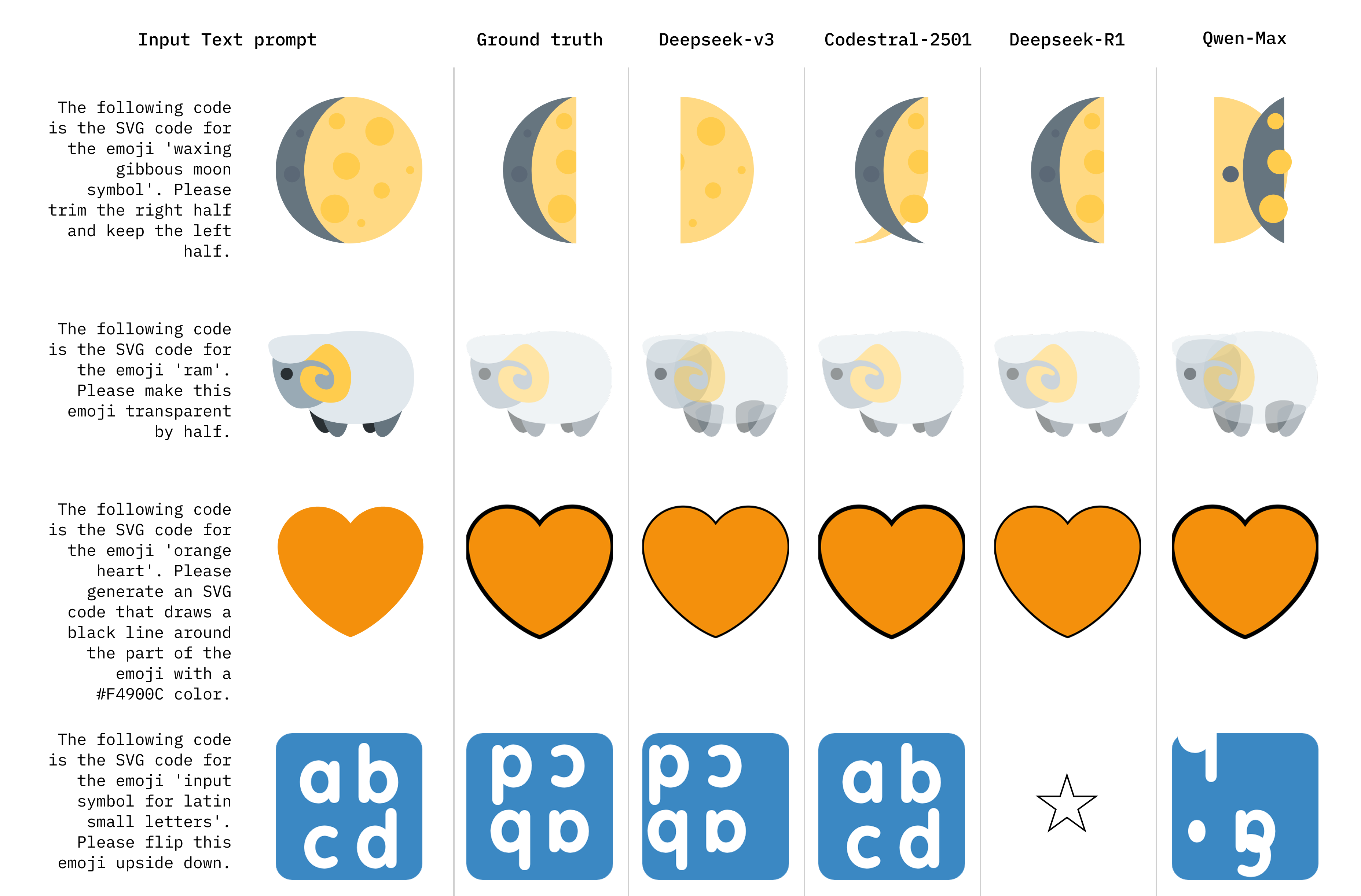}
    \caption{A comparison of SVGEditBench tasks for various models. 
    The symbol $\star$ indicates that rendered result is empty, because of incorrect transformation used by Deepseek-R1.}
    \label{fig:editing}
\end{figure}

To provide a visual comparison of generative abilities, we present randomly selected images produced by the best model from each group in Figure~\ref{fig:models_generation}. 
The results for the editing task are displayed in Figure~\ref{fig:editing}. 
These examples illustrate general differences in output quality and style across the models. 
Since IconShop was trained exclusively on black-and-white images, its outputs are limited, and it also struggles with processing long contextual inputs. 
Deepseek‑R1 demonstrates significantly enhanced quality compared to its no‑reasoning variant, particularly in its ability to refine details.

\begin{figure}
    \centering
    \includegraphics[width=0.9\linewidth]{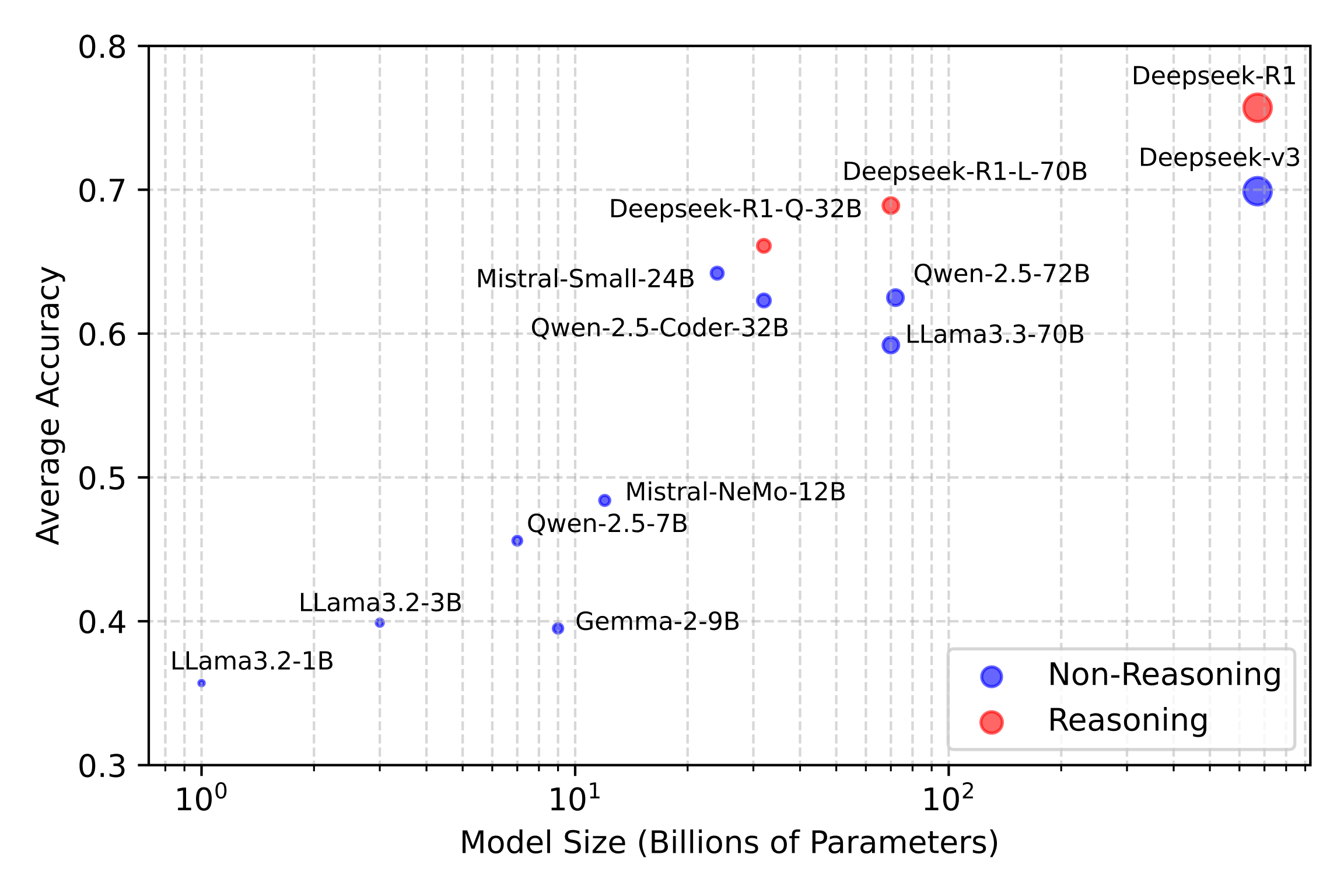}
    \caption{Comparison of reasoning models (and reasoning-based distillations) and non-reasoning models.}
    \label{fig:reason_with_no}
\end{figure}

Based on the results, we can conclude that, on average, reasoning improves performance and scores for SVG tasks. 
Even distilled versions of reasoning-based models outperform other models of the same size. 
Surprisingly, Codestral model obtained extremely high score in Editing. 
Figure~\ref{fig:reason_with_no} presents a comparison of all open-weight models listed in the table.

\section{Conclusion}\label{sec:conclusion}
Utilizing large language models for vector graphics processing is a promising area of research that remains largely unexplored.
Our study systematically reviews and categorizes existing LLM-based approaches for SVG generation, editing, and understanding. 
Through an extensive review and evaluation of $21$ models and $5$ benchmarks, we identify key trends and challenges in the field.

Our experimental results indicate that reasoning-enhanced LLMs significantly outperform their non-reasoning counterparts across multiple SVG-related tasks. 
Models such as DeepSeek-R1 demonstrate particularly strong capabilities in both generation and understanding tasks, while specialized models like IconShop remain limited by the simplicity of their training data. 
The analysis of benchmarks further highlights the lack of a comprehensive dataset covering all aspects of SVG processing, underscoring the need for future efforts in dataset creation and refinement.

Despite recent progress, several challenges remain. 
Many current approaches suffer from artifacts in complex and detailed image generation (even for images with $10$ shapes). 
Moreover, the models have limited ability to edit SVG structures intuitively, and generate inconsistencies during fine-grained vector manipulation. 
Additionally, access to high-quality, large-scale annotated datasets is crucial for further advancements in this field. 
Addressing these limitations will require novel model training strategies, innovations in LLM architecture, and data curation.

We anticipate that future research will focus on improving model robustness, enhancing SVG-specific tokenization strategies, and leveraging multi-modal approaches to bridge the gap between text-based LLM capabilities and the intricate requirements of vector graphics. 
By refining these methodologies, LLMs have the potential to revolutionize SVG generation, making vector images more accessible, efficient, editable, and customizable for designers, developers, and others. 
We also plan to analyze correlation between model performance and specific SVG complexity metrics and to discover how the relatively simple nature of benchmark SVGs (with average of 5-6 paths) limits applicability to complex real-world graphics.



%

\section*{Acknowledgments}
This paper was carried out as part of ITMO University project No. 624125 ``Development of advanced machine learning methods and algorithms''.


\bibliographystyle{amsplain}
{\small
\bibliography{main}}

\end{document}